\documentclass[runningheads]{llncs}
\usepackage{graphicx}
\usepackage{hyperref}
\usepackage{url}    
\usepackage{booktabs}       
\usepackage{amsfonts}
\usepackage{xcolor}
\usepackage{blkarray}
\usepackage{amsmath}
\usepackage{amssymb}
\usepackage{algorithm}
\usepackage{algpseudocode}
\usepackage{marvosym}
\renewcommand{\vec}[1]{\boldsymbol{#1}}

\begin{document}
\title{RED CoMETS: An ensemble classifier for symbolically represented multivariate time series}
\titlerunning{RED CoMETS}
%
\author{Luca A. Bennett\inst{1,2}\thanks{This author was with the University of Bristol while this research was undertaken but is currently affiliated with Awerian.}\textsuperscript{(\Letter)} \and Zahraa S. Abdallah\inst{1}\orcidID{0000-0002-1291-2918}}

\authorrunning{L. A. Bennett and Z. S. Abdallah}

\institute{
School of Engineering Mathematics and Technology\\ University of Bristol, Bristol, UK 
\\\email{zahraa.abdallah@bristol.ac.uk} \and 
Awerian, Cambridge, UK\\
\email{luca.bennett@awerian.net}\\\
}

\maketitle              
\begin{abstract}

Multivariate time series classification is a rapidly growing research field with practical applications in finance, healthcare, engineering, and more. The complexity of classifying multivariate time series data arises from its high dimensionality, temporal dependencies, and varying lengths. This paper introduces a novel ensemble classifier called RED CoMETS (Random Enhanced Co-eye for Multivariate Time Series), which addresses these challenges. RED CoMETS builds upon the success of Co-eye, an ensemble classifier specifically designed for symbolically represented univariate time series, and extends its capabilities to handle multivariate data. The performance of RED CoMETS is evaluated on benchmark datasets from the UCR archive, where it demonstrates competitive accuracy when compared to state-of-the-art techniques in multivariate settings. Notably, it achieves the highest reported accuracy in the literature for the `HandMovementDirection' dataset. Moreover, the proposed method significantly reduces computation time compared to Co-eye, making it an efficient and effective choice for multivariate time series classification.

\end{abstract}
\section{Introduction}
\label{sec: intro}

Problems involving the classification of time series data play a crucial role in various domains, including the sciences, data mining, finance, and signal processing. Time series and their classifiers can be categorised into two types: univariate and multivariate. Despite multivariate time series classification problems being more prevalent in real-world scenarios, the literature has historically focused more on the univariate case \cite{Ruiz2021TheAdvances}. Although recent studies have proposed promising methods to address multivariate time series classification \cite{Ruiz2021TheAdvances}, there still exists a gap, emphasising the need for accurate and efficient algorithms in this domain.

Traditional time series classifiers typically seek discriminatory features within the time series or adopt a holistic view of the entire series \cite{Bagnall2017TheAdvances}. They often concentrate on a single representation aspect, such as shape or frequency  \cite{Dempster2020ROCKET:Kernels}. However, time series classification problems can greatly differ in terms of training and testing sizes, dimensions, classes, series length, and class distribution. Consequently, a single approach cannot effectively handle all types of time series.

In this paper, we extend the techniques introduced by Co-eye for univariate time series classification \cite{Abdallah2020Co-eye:Series}, which draws inspiration from the compound eyes of insects. Co-eye utilizes two symbolic representation transformations, namely Symbolic Aggregate Approximation (SAX) \cite{Lin2007ExperiencingSeries} and Symbolic Fourier Approximation (SFA) \cite{Schafer2012SFA:Datasets}, to extract discriminatory features from the time series. These transformations generate multiple ``lenses" that can detect discriminatory features at various levels of granularity, capturing both fine details and broad shapes. By forming an ensemble of these lenses, Co-eye integrates different perspectives from the time and frequency domains, allowing for effective feature extraction in time series classification problems with diverse characteristics.

We propose a novel ensemble classifier for multivariate time series classification that builds upon Co-eye in two significant ways. Firstly, we enhance Co-eye's success in handling univariate problems and propose an improved approach that significantly reduces computation time without sacrificing accuracy. Secondly, we leverage this enhanced univariate approach as a foundation for a novel multivariate classifier, exploring two distinct techniques. Our proposed multivariate classifier is named RED CoMETS, which stands for Random Enhanced Co-eye for Multivariate Time Series. We evaluate RED CoMETS against state-of-the-art classifiers using datasets from the UCR archive \cite{BagnallUEARepository}, and it achieves state-of-the-art results.

The remainder of this paper is organised as follows: Section \ref{sec: related work} discusses relevant prior research. Section \ref{sec: optimise} provides details on our optimized univariate foundation built upon Co-eye. Section \ref{sec: multivariate} outlines the proposed extensions for multivariate classification. Section \ref{sec: experiments} presents the experimental results, specifically focusing on test accuracy. Finally, Section \ref{sec: conclusion} concludes the paper.

\section{Related Work}
\label{sec: related work}
Co-eye leverages the Symbolic Aggregate Approximation (SAX) \cite{Lin2007ExperiencingSeries} and Symbolic Fourier Approximation (SFA) \cite{Schafer2012SFA:Datasets} techniques to construct lenses, each offering a distinct view of the time series data in both the time and frequency domains. These lenses, represented by triplets denoted as $<s,\alpha,w>$, where $s$ indicates the choice between SAX and SFA, and $\alpha$ and $w$ are the hyperparameters for alphabet size and word length, respectively, provide Co-eye with a multi-resolution perspective \cite{Abdallah2020Co-eye:Series}. Through a careful ``pair selection" process, Co-eye identifies the most effective set of lenses for a given classification problem. During the classification phase, Co-eye builds a Random Forest \cite{TinKamHo1995RandomForests} for each lens using the transformed time series. These Random Forests' outputs are combined using a dynamic voting method, allowing the most confident lenses to be matched to specific sequences and effectively extracting discriminatory features \cite{Abdallah2020Co-eye:Series}. Co-eye has demonstrated competitive accuracies compared to state-of-the-art univariate classifiers when evaluated on datasets from the UCR archive \cite{Abdallah2020Co-eye:Series}.

The reviews by Bagnall et al. \cite{Bagnall2017TheAdvances} and Ruiz et al. \cite{Ruiz2021TheAdvances} provide a comprehensive overview of the strengths and weaknesses of different approaches, highlighting their performance on a range of datasets. This information is crucial in understanding the landscape of existing classifiers and identifying gaps or areas where further improvements can be made.

Dynamic Time Warping (DTW) \cite{Keogh2000ScalingApplications} is chosen as one of the benchmark classifiers. DTW utilizes a unique distance metric in combination with the 1-nearest neighbour classifier and serves as a baseline performance measure for ``good" time series classifiers. It was used as a benchmark by both Bagnall et al. \cite{Bagnall2017TheAdvances} and Ruiz et al.\cite{Ruiz2021TheAdvances}, making it a compelling target to surpass.

Another benchmark classifier is the Multiple Representation Sequence Learner (MrSEQL) \cite{LeNguyen2019InterpretableRepresentations}, which transforms time series into various symbolic representations and forms an ensemble using a SEQL classifier. While MrSEQL shares similarities with Co-eye in methodology, differences lie in the base classifier, parameterisation of symbolic representations, and voting methods \cite{LeNguyen2019InterpretableRepresentations}.

ROCKET (Random Convolutional Kernel Transform) \cite{Dempster2020ROCKET:Kernels} is a powerful classifier that has demonstrated exceptional performance in both univariate and multivariate time series classification. ROCKET leverages random convolutional kernels to transform time series data and apply a linear classifier to make predictions. It has achieved leading accuracies across the univariate UCR archive datasets while maintaining an extremely low computation time. The effectiveness and efficiency of ROCKET make it a natural choice to benchmark against for state-of-the-art performance.

HIVE-COTE (Hierarchical Vote Collective of Transformation-based Ensembles) \cite{hcote} is a heterogeneous ensemble classifier that combines multiple transformation based models. Its latest edition, HIVE-COTE 2.0 \cite{Middlehurst2021}, is currently the best-ranked multivariate time series classifier in terms of accuracy. HIVE-COTE constructs an ensemble of diverse classifiers, including shapelet-based, interval-based, and dictionary-based classifiers, and employs a hierarchical voting strategy to make predictions. The hierarchical nature of HIVE-COTE allows it to capture different levels of temporal patterns and achieve robust performance on a wide range of time series datasets. As the leading multivariate time series classifier, HIVE-COTE serves as the ``method to beat" for RED CoMETS.

In the realm of deep learning-based approaches for multivariate time series classification, InceptionTime \cite{IsmailFawaz2020InceptionTime:Classification} stands out. It is an ensemble of convolutional neural networks specifically designed for time series classification. InceptionTime introduces the concept of inception modules, which consist of parallel convolutional layers with different filter sizes. This design allows the network to capture diverse temporal patterns at multiple resolutions. InceptionTime has been identified by Ruiz et al. \cite{Ruiz2021TheAdvances} as the leading deep learning-based approach for both univariate and multivariate time series classification. Their review demonstrated that InceptionTime achieved top-performing accuracy across various datasets and outperformed many traditional and state-of-the-art classifiers. Therefore, it serves as a strong baseline for comparing the performance of RED CoMETS against deep learning-based approaches.

In addition to InceptionTime, deep learning architectures such as Long Short-Term Memory (LSTM) networks and Convolutional Neural Networks (CNNs) have gained popularity in time series classification. LSTM networks, a type of recurrent neural network (RNN), are capable of capturing long-term dependencies in sequential data and have shown promising results for classifying both univariate and multivariate time series \cite{karim2019multivariate}.

CNNs, on the other hand, are primarily known for their success in computer vision tasks, but they have also been applied to time series classification with remarkable outcomes. In the context of time series, 1D CNNs are often employed to learn hierarchical representations of input sequences by convolving filters across different time steps. This allows them to automatically extract relevant local patterns and capture higher-level representations of the data \cite{bai2018empirical,wang2017time}.

Deep learning-based approaches offer the advantage of automatically learning relevant features from raw time series data, obviating the need for handcrafted feature engineering. However, they often require large amounts of training data and significant computational resources for model training and optimization. Additionally, the interpretability of deep learning models can be challenging due to their black-box nature.
\section{Univariate Foundation}
\label{sec: optimise}

As described in Section \ref{sec: intro}, we first build on the univariate classification techniques introduced by Co-eye to create a new univariate classifier as a foundation for our multivariate extensions. We adapt the learning process of Co-eye, but introduce a new pair selection method and propose three replacement voting mechanisms. 

\subsection{Pair Selection}
\label{sec: ps}

Co-eye adopts a meticulous process for selecting lenses, involving two grid searches over the $\alpha-w$ parameter space for SAX and SFA, respectively. To construct an effective ensemble, each $<s,\alpha,w>$ triplet undergoes cross-validation, and pairs within a 1\% margin of the highest cross-validation accuracy are chosen. However, performing an exhaustive search and cross-validation for every $<\alpha, w>$ pair can be computationally demanding, as highlighted by Abdallah and Gaber \cite{Abdallah2020Co-eye:Series}. To address this bottleneck, we adopt a different approach inspired by the work of Bergstra and Bengio \cite{BergstraJames2012RandomSF}. They suggest that random searches can yield comparable performance to grid searches for hyperparameter selection. Therefore, we incorporate random selection in our methodology.

In Co-eye, the number of pairs is not predetermined. When generating pairs randomly, it is essential to preselect the number of SAX and SFA pairs. To ensure a balanced perspective of the time series and avoid voting bias, we opt for an equal number of SAX and SFA pairs. The selection of pairs is proportional to the length of the time series, with $\lfloor p * l \rfloor$ pairs independently chosen for SAX and SFA. Here, $0<p\leq1$ represents the proportion of pairs, and $l$ denotes the length of the time series. To determine the parameter space for random selection, we draw pairs uniformly from the $\alpha-w$ space defined by Abdallah and Gaber \cite{Abdallah2020Co-eye:Series}. We evaluate four different values of $p$, namely 0.05, 0.1, 0.15, and 0.2, denoted as R5\%, R10\%, R15\%, and R20\%, respectively. These values enable us to explore the impact of different proportions of pairs on the ensemble construction process.

By adopting this approach, we aim to strike a balance between computational efficiency and lens selection effectiveness, ensuring that Co-eye can efficiently construct an ensemble of lenses while capturing diverse perspectives of the data.

\subsection{Voting}
\label{sec:voting}

To enhance accuracy and robustness, we propose three voting methods to replace Co-eye's existing dynamic voting approach. Let's consider Co-eye applied to a dataset with $n$ classes ${c_1, \dots, c_n}$ and $m$ samples. Each base Random Forest classifier generates an $m\times n$ matrix, denoted as:

\begin{equation}
M_i = 
\begin{blockarray}{cccc}
 & c_1 & \dots & c_n \\
\begin{block}{c(ccc)}
  \text{Sample } 1 \text{ } & P(c = c_1) & \dots & P(c = c_n) \\
  \vdots & \vdots & \ddots & \vdots \\
  \text{Sample } m \text{ } & P(c = c_1) & \dots & P(c = c_n) \\
\end{block}
\end{blockarray}
\text{ }\text{ .}
\end{equation}

Therefore, Co-eye produces a set of matrices, denoted as $S_M = {M_1, \dotsc, M_k}$, where $k$ represents the number of classifiers in the ensemble. Voting can be seen as a function on $S_M$, resulting in a vector of class labels for the $m$ samples. We introduce three new voting methods based on the sum rule (SR) scheme outlined in Algorithm \ref{alg:sum rule}, employing different weight generation functions.

\begin{algorithm}
\caption{Sum Rule Scheme}\label{alg:sum rule}
\begin{algorithmic}[1]
\Procedure{SumRule}{$S_M$}
\State $w \gets \text{getWeights()}$
\State $\text{weightedMats} \gets w*S_M$ \Comment Element-wise multiplication.
\State $\text{sum} \gets \sum_{k}\text{weightedMats}$ \Comment Element-wise addition.
\For{\text{row in sum}}
\State label $\gets$ max(row)
\EndFor
\State \Return{labels}
\EndProcedure
\end{algorithmic}
\end{algorithm}

The first voting method is the simplest, employing uniform weights of one across the ensemble. Although efficient, we hypothesize that a more sophisticated weighting scheme could yield better results. Intuitively, matrices with higher confidence in their predictions should carry more weight. Thus, matrices with greater row-wise maximum confidences can be considered to be more confident. For a matrix $M_i \in S_M$ with $m$ rows, the set of row-wise maxima can be defined as $R^{max}_{i} = {\text{row}{max}(j) \mid \forall j \in [m]}$, where $\text{row}{max}(j)$ represents the maximum value of row $j$ in matrix $M_i$, and $[m] = {1, \dots, m}$. Let $\overline{R^{max}_{i}}$ denote the mean of the row-wise maxima. Our second voting scheme then assigns weights as $\boldsymbol{w} = [\overline{R^{max}_{1}}, \dotsc, \overline{R^{max}_{k}}]$.

Instead of using $S_M$ directly for weight generation, Large et al. \cite{Large2019AEstimates} demonstrated the effectiveness of weights determined through cross-validation. Hence, our third proposed voting method is as follows: A Random Forest is built for each $<s, \alpha, w>$ triplet, and accuracy is calculated using 5-fold cross-validation, a value supported by Burman \cite{Burman1989AMethods}. The cross-validation accuracies are then used as weights for their respective matrices. Note that this method is significantly more computationally expensive than the other two approaches. However, unlike Co-eye's pair selection process, cross-validation is applied only to selected triplets rather than the entire $\alpha-w$ parameter space, making it computationally viable.

We refer to the three voting methods as SR Uniform, SR Mean-Max, and SR Validation, respectively.

\section{Developing RED CoMETS}
\label{sec: multivariate}

We anticipate that extending the multi-resolution perspectives of Co-eye, which is effective for univariate time series classification using the time and frequency domains, will be equally successful for multivariate datasets. In the literature, both forests \cite{ARForests} and symbolic representations \cite{mvSymbR} have achieved favourable results in this regard. To enable univariate classifiers to handle multivariate time series, we present two approaches. When combined with the univariate foundation established from Co-eye in Section \ref{sec: optimise}, these approaches form RED CoMETS (Random Enhanced Co-eye for Multivariate Time Series).

\subsection{Concatenating Approach}
\label{sec: concat}

One intuitive approach to address multivariate time series classification is to reduce it to the more extensively studied univariate case. This can be achieved by sequentially concatenating the dimensions of a multivariate dataset. For a multivariate time series with a length of $n$ and $d$ dimensions, this method generates a univariate time series of length $nd$. Algorithm \ref{alg: glue dim coeye} demonstrates the application of this method to our univariate foundation. When utilizing the random pair selection technique described in Section \ref{sec: ps}, the number of lenses is proportional to the length of the time series. However, for computational efficiency, it was decided that if random pair selection is used, the number of lenses will be determined based on the length of the time series before concatenation, i.e., proportional to $n$ rather than $nd$.

\begin{algorithm}
\caption{Concatenating Approach}\label{alg: glue dim coeye}
\begin{algorithmic}[1]
\Procedure{ConcatenatingApproach}{TS} \Comment{TS is a multivariate dataset}
\For{dimension $\in$ TS}
\State append(concatTS, dimension)
\EndFor
\State \Return UnivariateFoundation(concatTS)
\EndProcedure
\end{algorithmic}
\end{algorithm}

\subsection{Ensembling Approach}
\label{sec: ensemble}

Another approach to handling multivariate datasets is to construct an ensemble over the dimensions. This method, recommended by Ruiz et al. \cite{Ruiz2021TheAdvances}, involves building a univariate classifier for each dimension and combining their predictions for the overall classification.

Since our univariate foundation is an ensemble classifier, this leads to an ensemble of ensembles. Consequently, there are two sub-approaches depending on how the ensemble results are combined. Algorithms \ref{alg: ensemb dim app1} and \ref{alg: ensemb dim app2} outline these sub-approaches. Approach 1 combines the set of matrices, $S_M$, produced by each base classifier into a single superset $S_\text{all} = S_{M1} \cup S_{M2} \cup \dots S_{Md}$, where $S_{Mi}$ represents the set of matrices returned for the $i$th dimension. Voting is then applied as usual to $S_\text{all}$. Approach 2 performs voting in two stages. For each dimension, $S_{Mi}$ is fused into a single matrix, $F_i$, using one of the sum rule methods outlined in Section \ref{sec:voting}. For the $i$th dimension, $F_i = \sum_{k} \vec{w}S_{Mi}$, where $\vec{w}$ is a vector of weights. Subsequently, a second round of voting is applied to the set of fused matrices across all dimensions, denoted as $S_F = \{F_i\mid \forall i \in [d]\}$, where $[d] = {1,\dotsc,d}$, resulting in the final classification. Different voting methods can be employed for the fusion and final classification stages.

\begin{algorithm}
\caption{Ensembling Approach 1}\label{alg: ensemb dim app1}
\begin{algorithmic}[1]
\Procedure{EnsemblingApproach1}{TS} \Comment{TS is a multivariate dataset}
\For{dimension $\in$ TS}
\State $S_M \gets$ UnivariateFoundation(dimension)
\State append($S_\text{all}, S_M$)
\EndFor
\State \Return{vote($S_\text{all}$)}
\EndProcedure
\end{algorithmic}
\end{algorithm}

\begin{algorithm}
\caption{Ensembling Approach 2}\label{alg: ensemb dim app2}
\begin{algorithmic}[1]
\Procedure{EnsemblingApproach2}{TS} \Comment{TS is a multivariate dataset}
\For{dimension $\in$ TS}
\State $S_M \gets$ UnivariateFoundation(dimension)
\State $F_i \gets \sum_k w*S_M$ \Comment{Element-wise operations}
\State append($S_F, F_i$)
\EndFor
\State \Return{vote($S_F$)}
\EndProcedure
\end{algorithmic}
\end{algorithm}

\subsection{RED CoMETS}
The univariate foundation described in Section \ref{sec: optimise}, which builds upon the innovative time series classification approach introduced by Co-eye \cite{Abdallah2020Co-eye:Series}, incorporates a new random pair selection process and three new voting methods. By combining the two proposed multivariate extensions from Sections \ref{sec: concat} and \ref{sec: ensemble} with our univariate foundation, we establish a novel multivariate classifier (RED CoMETS).

\section{Experiments and Evaluation}
\label{sec: experiments}
We evaluate our univariate foundation and RED CoMETS on univariate and multivariate datasets respectively from the UCR archive. We demonstrate that our univariate foundation is more accurate and approximately 40 times faster than Co-eye. RED CoMETS is shown to achieve accuracies comparable to the state-of-the-art classifiers outlined in Section \ref{sec: related work}. Our code and full results are available on GitHub \footnote{\url{https://github.com/zy18811/RED-CoMETS}}.

\subsection{Experimental Design}
All of our experiments were conducted with the 111 datasets from the UCR archive \cite{BagnallUEARepository} used by Bagnall et al. \cite{Bagnall2017TheAdvances} and Ruiz et al. \cite{Ruiz2021TheAdvances} in their reviews, consisting of 85 univariate and 26 multivariate datasets. This allows for comparison to the results recorded by Bagnall et al. \cite{Bagnall2017TheAdvances} and Ruiz et al. \cite{Ruiz2021TheAdvances} in their reviews of state-of-the-art classifiers. For consistency and to allow direct comparison, our results show the average over 30 trials on each data using 30 stratified resamples. Each resample is seeded by its sample number, such that each classifier is evaluated on identical samples and results are reproducible. Note that both SAX and SFA z-normalise the time series as their initial step. For the multivariate datasets, this means that the concatenating approach normalises the joint time series while the ensembling approach normalises each dimension independently.

We produce results for Co-eye, our univariate foundation, and RED CoMETS. Results for DTW and univariate ROCKET were taken from Bagnall et al. \cite{Bagnall2017TheAdvances} and Dempster et al. \cite{Dempster2020ROCKET:Kernels} respectively. The results for DTW$_D$, MrSEQL, InceptionTime, and multivariate ROCKET were taken from Ruiz et al. \cite{Ruiz2021TheAdvances}. The results for HIVE COTE-2.0 were taken from the author's website \cite{BagnallUEARepository}. The default accuracy for predicting the majority class is also included and is taken from Bagnall et al. \cite{Bagnall2017TheAdvances} and Ruiz et al. \cite{Ruiz2021TheAdvances} for the univariate and multivariate datasets respectively. The voting methods proposed in Section \ref{sec:voting} were evaluated using the R5\% pair selection described in Section \ref{sec: ps} to minimise computation time.

To compare multiple classifiers over multiple datasets, critical difference (CD) diagrams are used \cite{Demsar2006StatisticalSets}. Current literature \cite{Benavoli2016ShouldMean-Ranks} suggests abandoning the post hoc test originally suggested by Dem\v{s}ar \cite{Demsar2006StatisticalSets}, instead forming cliques using pairwise tests, with the Holm correction being made in the case of multiple testing. The classifiers are first ranked using the Friedman test, then grouped into cliques using pairwise Wilcoxon signed rank tests with the Holm adjustment \cite{Bagnall2017TheAdvances,Ruiz2021TheAdvances}. Cliques represent groups of classifiers between which there is no statistically significant pairwise difference. A Python implementation produced by Fawaz et al. \cite{IsmailFawaz2019DeepReview} was used to create the CD diagrams presented in this paper.

\subsection{Univariate Foundation}
\label{sec: opt results}
\subsubsection{Pair Selection}
\label{sec: ps results}
\begin{figure}
    \centering
    \includegraphics[width = 0.8\textwidth]{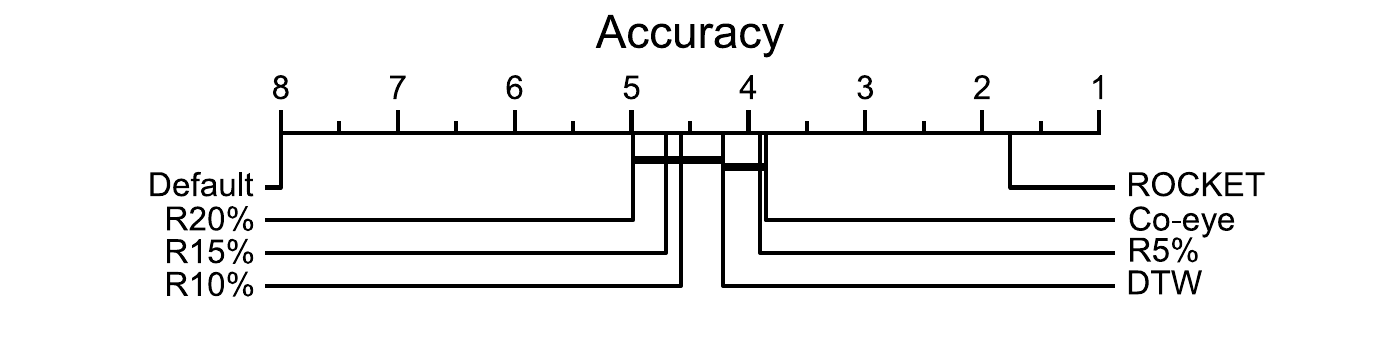}
    \caption{Test accuracy critical difference diagram for random pair selection methods against Co-eye averaged over 30 resamples for each of the 85 univariate UCR datasets. Default accuracy, DTW, and ROCKET are included as benchmarks.}
    \label{fig: pair selection cd diag}
\end{figure}

The four random pair selection methods outlined in Section \ref{sec: ps} were evaluated on the 85 univariate datasets from the UCR archive in order to evaluate their effectiveness against Co-eye. Default accuracy, DTW, and univariate ROCKET are included as benchmarks. Figure \ref{fig: pair selection cd diag} shows the test accuracy critical difference (CD) diagram for the pair selection methods. It can be seen that there are two distinct cliques containing R10\%, R15\%, and R20\% and Co-eye and R5\% respectively, with DTW found in both. Both cliques outperformed default accuracy with statistical significance. ROCKET significantly outperformed all others. R10\%, R15\%, and R20\% all performed worse in terms of accuracy than Co-eye, and are removed from contention. There is no statistically significant pairwise difference in test accuracy between R5\% and Co-eye. 

\begin{figure}[ht!]
    \centering
    \includegraphics[width = 0.8\textwidth]{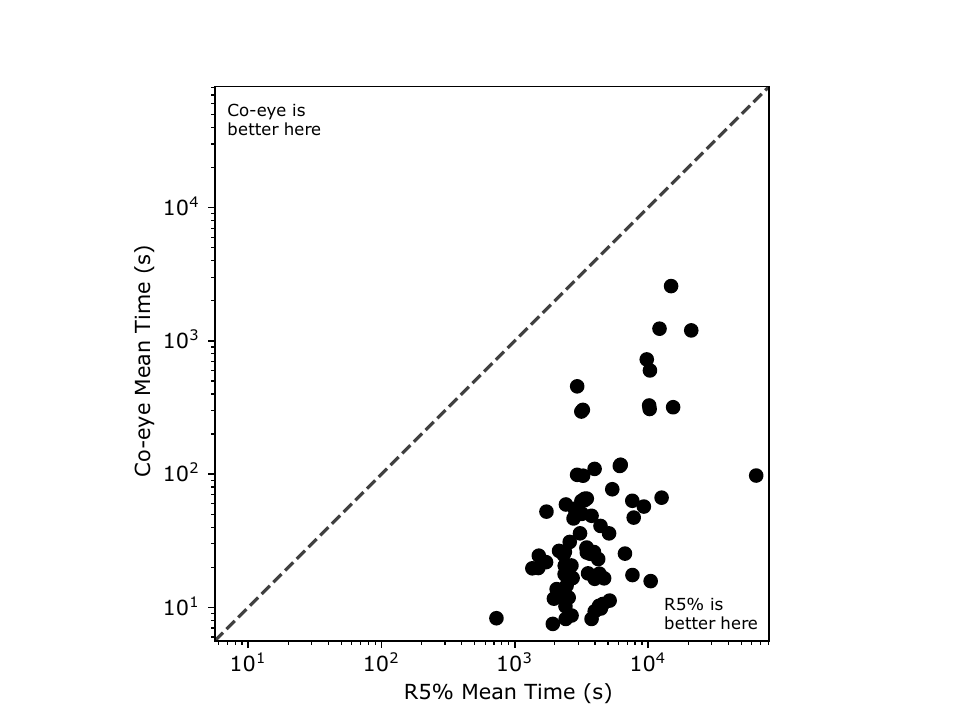}
    \caption{Pairwise comparison  of total mean train and test time between Co-eye and R5\% averaged over 30 stratified resamples of the 85 univariate UCR datasets.}
    \label{fig: timing comp}
\end{figure}

Figure \ref{fig: timing comp} shows a pairwise comparison of mean train and test time between Co-eye and R5\% on the 85 univariate UCR datasets. It can be seen that R5\% is significantly faster than Co-eye in all cases, averaging approximately 40 times faster over the 85 datasets. As such, R5\% is a pronounced improvement over Co-eye: 40 times faster with no statistically significant difference in test accuracy. For R5\%, Kendall's $\tau$ coefficient was calculated between characteristics of each dataset and the associated total train and test time, with values of 0.41, 0.42, 0.33, and 0.78 for train size, test size, number of classes, and series length respectively. As one would expect, there is a positive correlation for all values, with series length as the most significant determinant of train and test time.

\subsubsection{Voting}
\label{sec: voting res}
\begin{figure}[ht!]
    \centering
    \includegraphics[width = 0.8\textwidth]{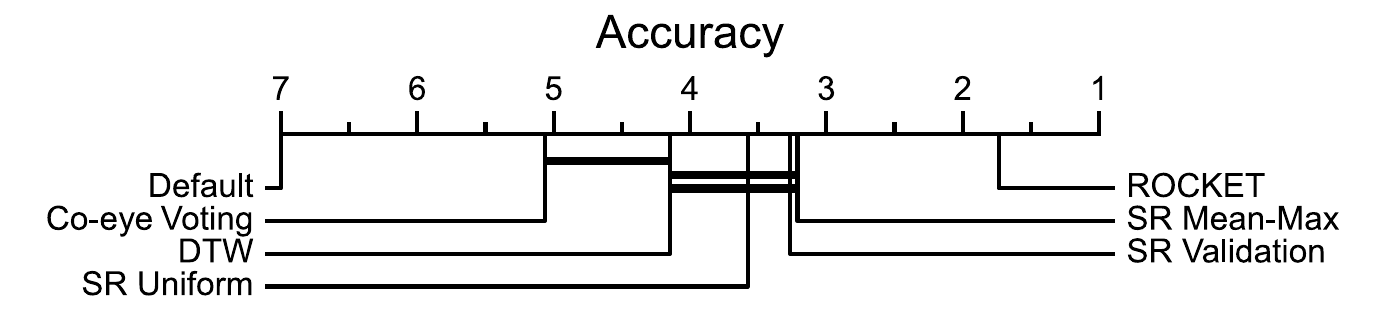}
    \caption{Test accuracy critical difference diagram for proposed voting methods against Co-eye averaged over 30 resamples for each of the 85 univariate UCR datasets. Default accuracy, DTW, and ROCKET are included as benchmarks.}
    \label{fig: voting cd diag}
\end{figure}

Section \ref{sec:voting} proposed three voting methods, aiming to outperform the dynamic voting method used by Co-eye in terms of test accuracy. As done above for pair selection, the voting methods were evaluated on the 85 univariate datasets from the UCR archive with default accuracy, DTW, and univariate ROCKET as benchmarks. Figure \ref{fig: voting cd diag} shows the test accuracy CD diagram for the voting methods. It can be seen that the three proposed voting methods all performed better than Co-eye's dynamic voting method with statistical significance. The three voting methods are cliqued, indicating no significant pairwise difference between them. As such, all three voting methods are taken forward for evaluation as part of RED CoMETS. 

\subsection{RED CoMETS}
\label{sec: mv coeye res}

 There are nine variants of RED CoMETS, which result from different combinations of a voting method and the multivariate extension. These variants are referred to by the names presented in Table \ref{tab: names}. It is worth noting that the validation voting method is not utilised with the ensembling dimensions multivariate extension due to initial experiments demonstrating computational infeasibility.

\renewcommand{\arraystretch}{.92}
\begin{table}[ht!]
\caption{RED CoMETS variants.}
\label{tab: names}
\centering
\begin{tabular}{@{}llcll@{}}
\toprule
Name         & Approach  & \multicolumn{1}{l}{Sub-Approach} & Voting Method 1 & Voting Method 2 \\ \midrule
RED CoMETS-1 & Concatenating    & n/a                          & Uniform         & n/a             \\
RED CoMETS-2 & Concatenating   & n/a                          & Mean-Max        & n/a             \\
RED CoMETS-3 & Concatenating    & n/a                          & Validation      & n/a             \\
RED CoMETS-4 & Ensembling & 1                            & Uniform         & n/a             \\
RED CoMETS-5 & Ensembling & 1                            & Mean-Max        & n/a             \\
RED CoMETS-6 & Ensembling & 2                            & Uniform         & Uniform         \\
RED CoMETS-7 & Ensembling & 2                            & Uniform         & Mean-Max        \\
RED CoMETS-8 & Ensembling & 2                            & Mean-Max        & Mean-Max        \\
RED CoMETS-9 & Ensembling & 2                            & Mean-Max        & Uniform         \\ \bottomrule
\end{tabular}
\end{table}

We evaluate the RED CoMETS variants against each other and the multivariate benchmarks discussed in Section \ref{sec: related work} (DTW$_D$, MrSEQL, InceptionTime, ROCKET, and HIVE-COTE 2.0). When evaluated by Ruiz et al. \cite{Ruiz2021TheAdvances}, InceptionTime and MrSEQL were unable to complete all 26 datasets, with InceptionTime failing on `EigenWorms' due to memory errors and MrSEQL failing to complete `FaceDetection' and `PhonemeSpectra' within the set time constraints. Likewise, all variants of RED CoMETS were unable to complete `Eigenworms'. As such, results for these datasets will not be used in our comparison, leaving 23 datasets for evaluation. Based on the results shown in Section \ref{sec: ps results}, all variants of RED CoMETS are evaluated using the R5\% pair selection method. As Section \ref{sec: voting res} demonstrated no statistically significant difference between the three proposed voting methods, all nine RED CoMETS variants shown in Table \ref{tab: names} are evaluated. 

\begin{figure}
    \centering
    \includegraphics[width = 0.8\textwidth]{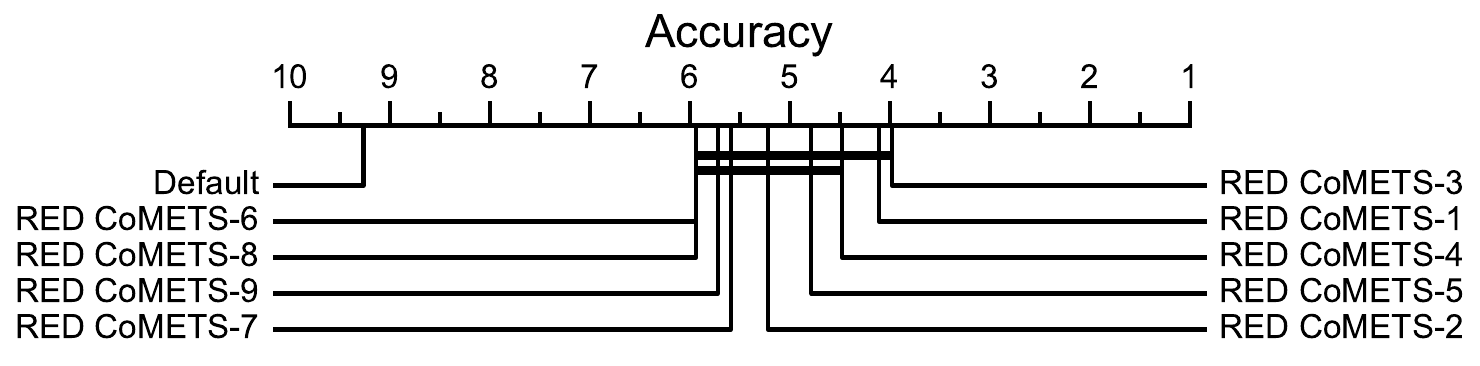}
    \caption{Test accuracy critical difference diagram for RED CoMETS variants averaged over the 23 UCR datasets.}
    \label{fig: rc comp cd diag}
\end{figure}

We first analyse the nine variants of RED CoMETS. It can be seen in Figure \ref{fig: rc comp cd diag} that there is no statistically significant pairwise difference in test accuracy between the nine RED CoMETS variants, with the default accuracy being outperformed with statistical significance in all cases. However, looking at the results shown in Table \ref{tab: rc results}, RED CoMETS-3 has both the highest mean accuracy and number of wins, indicating that it is both the most accurate and most reliable of the nine RED CoMETS variants.

\newcommand{\hlight}[1]{\underline{\textbf{#1}}}
\begin{table}[!ht]
\caption{Summary of RED CoMETS results showing mean accuracy across 30 resamples for each variant and multivariate dataset. The mean and number of wins are also shown. The greatest values on each row are shown in underlined bold.}
\label{tab: rc results}
\centering
\begin{tabular}{@{}llllllllll@{}}
\cmidrule(l){2-10}
 & \multicolumn{9}{c}{RED CoMETS-\textless{}N\textgreater{} (\%)} \\ \midrule
Dataset & \multicolumn{1}{c}{1} & \multicolumn{1}{c}{2} & \multicolumn{1}{c}{3} & \multicolumn{1}{c}{4} & \multicolumn{1}{c}{5} & \multicolumn{1}{c}{6} & \multicolumn{1}{c}{7} & \multicolumn{1}{c}{8} & \multicolumn{1}{c}{9} \\ \midrule
AWR & \hlight{97.73} & 97.60 & \hlight{97.73} & 96.22 & 96.02 & 95.20 & 95.16 & 94.91 & 94.18 \\
AF & 30.00 & 29.11 & 29.78 & 28.89 & 28.89 & \hlight{32.00} & \hlight{32.00} & 31.33 & 31.33 \\
BM & \hlight{98.17} & 98.00 & \hlight{98.17} & 79.58 & 79.42 & 81.67 & 81.92 & 81.75 & 82.08 \\
CR & 97.08 & \hlight{97.13} & \hlight{97.13} & 92.59 & 92.73 & 89.31 & 89.31 & 89.58 & 89.68 \\
DDG & 59.60 & 54.47 & \hlight{62.27} & 20.53 & 19.13 & 20.47 & 20.53 & 19.60 & 19.40 \\
EP & 85.14 & 83.60 & \hlight{85.29} & 60.31 & 53.26 & 58.36 & 57.83 & 51.79 & 50.46 \\
ER & \hlight{93.54} & 92.38 & 93.51 & 91.23 & 91.19 & 85.68 & 85.63 & 85.88 & 85.79 \\
EC & 27.55 & 27.60 & 27.59 & 33.13 & \hlight{33.36} & 32.60 & 32.56 & 32.69 & 32.53 \\
FM & 51.93 & 50.30 & 51.60 & 52.20 & \hlight{52.53} & 52.10 & 52.10 & 52.40 & 52.43 \\
HMD & 54.20 & 54.57 & \hlight{55.30} & 44.36 & 44.68 & 42.40 & 42.40 & 42.81 & 42.99 \\
HW & \hlight{32.73} & 31.67 & 32.60 & 28.97 & 29.05 & 27.33 & 27.32 & 27.64 & 27.68 \\
HB & 66.44 & 65.38 & 66.50 & 71.02 & 70.98 & 71.02 & \hlight{71.12} & 70.98 & 71.04 \\
LIB & \hlight{78.33} & 75.93 & \hlight{78.33} & 73.33 & 72.89 & 58.85 & 58.85 & 57.85 & 57.85 \\
LSST & 15.96 & 05.76 & \hlight{50.93} & 08.90 & 08.07 & 05.35 & 05.21 & 04.03 & 03.71 \\
MI & 51.00 & 51.20 & 50.97 & 51.33 & 51.50 & 51.37 & 51.40 & \hlight{51.57} & 51.53 \\
NATO & 82.04 & 81.81 & \hlight{82.30} & 73.54 & 73.78 & 72.41 & 72.83 & 72.15 & 72.72 \\
PEMS & 78.30 & 77.59 & 78.30 & 90.98 & 91.89 & 92.08 & 92.49 & 93.14 & \hlight{93.66} \\
PD & 88.00 & 82.16 & \hlight{88.17} & 76.32 & 76.24 & 63.64 & 63.64 & 64.21 & 64.21 \\
RS & \hlight{83.05} & 72.74 & 82.87 & 78.60 & 78.82 & 75.46 & 75.70 & 75.61 & 75.77 \\
SRS1 & 85.46 & 85.51 & 85.46 & 86.47 & \hlight{86.50} & 86.35 & 86.36 & 86.38 & 86.41 \\
SRS2 & 51.89 & 52.02 & 52.00 & \hlight{52.39} & 52.35 & 52.37 & 52.37 & 52.35 & 52.33 \\
SWJ & 38.89 & 38.44 & 38.44 & 43.33 & 43.33 & 44.67 & \hlight{44.89} & 44.22 & 44.44 \\
UW & \hlight{88.61} & 88.53 & 88.60 & 84.20 & 84.14 & 81.09 & 80.99 & 80.95 & 80.83 \\ \midrule
Mean & 66.77 & 64.94 & \hlight{68.43} & 61.67 & 61.34 & 59.64 & 59.68 & 59.30 & 59.26 \\ \midrule
Wins & 5.5 & 0.5 & \hlight{8} & 1 & 3 & 0.5 & 2.5 & 1 & 1 \\ \bottomrule
\end{tabular}
\end{table}

\begin{figure}
    \centering
    \includegraphics[width = 0.8\textwidth]{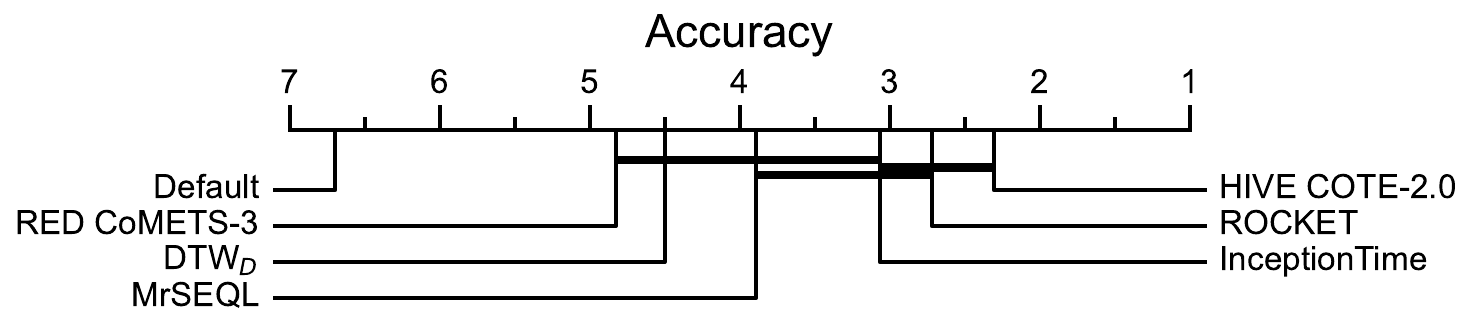}
    \caption{Test accuracy critical difference diagram for RED CoMETS-3 against the state-of-the-art classifiers averaged over the 23 UCR datasets}
    \label{fig: mv sota comp cd diag}
\end{figure}

Having identified RED CoMETS-3 as the most effective variant, we now evaluate it against the state-of-the-art methods identified in Section \ref{sec: related work}. It can be seen from Figure \ref{fig: mv sota comp cd diag} that, excluding default accuracy, RED CoMETS-3 has the lowest ranking in terms of test accuracy. However, the cliques indicate that there is no statistically significant difference in accuracy between RED CoMETS-3 and DTW$_D$, MrSEQL, and InceptionTime, demonstrating that RED CoMETS-3 is competitive with state-of-the-art multivariate classifiers. 

We now further analyse the performance of RED CoMETS-3 in relation to the benchmarks, with Table \ref{tab: acc diffs} showing the differences in test accuracy. RED CoMETS-3 was able to beat all of the benchmarks on at least four of the datasets. Both the mean and median difference in accuracy between RED CoMETS-3 and DTW$_D$, MrSEQL, and InceptionTime is less than 5\%, concurring with Figure \ref{fig: mv sota comp cd diag}. Looking at the maxima and minima, it can be seen that RED CoMETS-3 greatly outperforms the benchmarks on some datasets and vice versa. In fact, RED CoMETS-3 consistently outperforms the state-of-the-art benchmarks on a small number of datasets, beating all of the benchmarks on HMD, four on AF and DDG, and three on ER, SRS1, and SRS2. In other words, just six datasets account for 22 out of the 28 wins shown in Table \ref{tab: acc diffs}. Five of these six datasets are categorised as EEG, ECG, or spectrographic. Hence, it is apparent that RED CoMETS attains its best performance on datasets with minimal phase shifting (this was also found to be the case for Co-eye by Abdallah and Gaber \cite{Abdallah2020Co-eye:Series}).

\begin{table}
\caption{Summary of the test accuracy differences between RED CoMETS-3 and the benchmarks for the multivariate UCR datasets. Negative is better for RED CoMETS-3.}
\label{tab: acc diffs}
\begin{tabular}{@{}llllllll@{}}
\toprule
Classifier & Mean (\%) & Median (\%) & Max (\%) & Min (\%) & STD (\%) & Wins & Losses \\ \midrule
DTW$_D$ & 0.68 & 1.69 & 28.60 & -24.98 & 10.32 & 9 & 14 \\
MrSEQL & 4.39 & 3.93 & 65.60 & -33.33 & 18.31 & 6 & 17 \\
InceptionTime & 3.64 & 2.63 & 64.51 & -65.59 & 21.98 & 5 & 18 \\
ROCKET & 5.09 & 5.25 & 24.06 & -16.13 & 8.66 & 4 & 19 \\
HIVE COTE-2.0 & 7.50 & 5.33 & 51.50 & -15.52 & 13.17 & 4 & 19 \\ \bottomrule
\end{tabular}
\end{table}

HIVE COTE-2.0 and ROCKET are considered the current best within the state-of-the-art as discussed in Section \ref{sec: related work}. Figure \ref{fig: mv sota comp cd diag} corroborates this, with them being ranked first and second respectively. We now compare RED CoMETS-3 against them in more detail, seeking to better understand the disparities shown in Table \ref{tab: acc diffs}. It can be seen from Table \ref{tab: mv sota} that HIVE-COTE 2.0 retains its place as the current best classifier in terms of test accuracy with both the greatest mean accuracy and number of wins. However, RED CoMETS-3 is still able to hold its own against ROCKET and HIVE-COTE 2.0, beating both of them on four of the datasets. Furthermore, the result obtained for the HMD dataset, 55.30\%, is greater than any reported in the literature \cite{BagnallUEARepository}, representing a notable improvement to the state-of-the-art. 

\begin{table}[!ht]
\caption{Results for ROCKET, HIVE-COTE 2.0, and RED CoMETS-3 showing mean test accuracy across 30 resamples of each multivariate dataset. The mean and number of wins are also shown. The greatest values on each row are shown in underlined bold.}
\label{tab: mv sota}
\centering
\begin{tabular}{@{}llll@{}}
\toprule
Dataset & \multicolumn{1}{c}{ROCKET (\%)} & \multicolumn{1}{c}{HIVE COTE-2.0 (\%)} & \multicolumn{1}{c}{RED CoMETS-3 (\%)} \\ \midrule
AWR & 99.56 & \hlight{99.58} & 97.73 \\
AF & 24.89 & 28.22 & \hlight{29.78} \\
BM & \hlight{99.00} & 98.92 & 98.17 \\
CR & \hlight{100.00} & 99.95 & 97.13 \\
DDG & 46.13 & 49.87 & \hlight{62.27} \\
EP & 99.08 & \hlight{99.83} & 85.29 \\
ER & 98.05 & \hlight{98.51} & 93.51 \\
EC & 44.68 & \hlight{79.09} & 27.59  \\
FM & \hlight{55.27} & 55.23 & 51.60 \\
HMD & 44.59 & 39.77 & \hlight{55.30} \\
HW & \hlight{56.67} & 56.34 & 32.60 \\
HB & 71.76 & \hlight{72.86} & 66.50 \\
LIB & 90.61 & \hlight{92.69} & 78.33 \\
LSST & 63.15 & \hlight{63.70} & 50.93 \\
MI & 53.13 & \hlight{53.17} & 50.97 \\
NATO & 88.54 & \hlight{89.20} & 82.30 \\
PEMS & \hlight{99.56} & \hlight{99.56} & 88.17 \\
PD & 85.63 & \hlight{99.81} & 78.30 \\
RS & 92.79 & \hlight{93.05} & 82.87 \\
SRS1 & 86.55 & \hlight{87.87} & 85.46 \\
SRS2 & 51.35 & 50.46 & \hlight{52.00} \\
SWJ & \hlight{45.56} & 43.78 & 38.44 \\
UW & 94.43 & \hlight{94.89} & 88.60 \\ \midrule
Mean & 73.52 & \hlight{75.93} & 68.43 \\ \midrule
Wins & 5.5 & \hlight{13.5} & 4 \\ \bottomrule
\end{tabular}
\end{table}

\section{Conclusion}
\label{sec: conclusion}
RED CoMETS is a novel ensemble classifier for multivariate time series that builds on the success of Co-eye. In order to build a univariate foundation for our classifier, we adapted Co-eye's use of multiple symbolic representations to gain a multi-resolution perspective of both the time and frequency domains. However, we introduced a random pair selection process in order to overcome the bottleneck in Co-eye \cite{Abdallah2020Co-eye:Series}. We also proposed and evaluated three new voting methods. Our adaption of Co-eye was extremely successful, achieving an approximately 40 times increase in speed and small but statistically significant gains in accuracy in comparison to Co-eye. 

Two multivariate extensions were then applied to our univariate classifier. The different possible combinations of the multivariate extensions and voting methods resulted in the nine variants of RED CoMETS shown in Table \ref{tab: names}. These were evaluated against state-of-the-art classifiers on 23 multivariate datasets from the UCR archive \cite{BagnallUEARepository}, following the methodology of Ruiz et al. \cite{Ruiz2021TheAdvances}. 

RED CoMETS-3 was identified as the clear best out of the nine variants in both accuracy and reliability and was demonstrated to have no statistically significant pairwise difference in accuracy to several of the state-of-the-art benchmarks. RED CoMETS-3 was able to outperform both ROCKET and HIVE COTE-2.0, the current best-in-class, on four of the 23 datasets and achieved an accuracy greater than reported by any classifier in the literature on the `HandMovementDirection' dataset. It was noted that RED CoMETS attains its best performance on datasets with no significant phase shifting.

There is room to further improve RED CoMETS-3 in both the R5\% pair selection and SR Validation voting method. For R5\%, a subset of the datasets could be used to learn the optimal bounds for the $\alpha-w$ parameter space, similar to the methodology used by Dempster et al. \cite{Dempster2020ROCKET:Kernels} when learning the kernel parameter space for ROCKET. SR Validation could be improved by emulating the scheme proposed by Large et al. \cite{Large2019AEstimates} in which the weights are raised to a power in order to amplify differences between base classifiers.

%
%
%

\bibliographystyle{splncs04}
\bibliography{references}
%




\end{document}